\documentclass[tinyml]{acmart}

\AtBeginDocument{%
  \providecommand\BibTeX{{%
    \normalfont B\kern-0.5em{\scshape i\kern-0.25em b}\kern-0.8em\TeX}}}

\setcopyright{rightsretained}
\copyrightyear{2022}
\acmYear{2022}

\usepackage{amsmath,amsfonts}
\usepackage{algorithm}
\usepackage{algpseudocode}
\usepackage{graphicx}
\usepackage{textcomp}
\usepackage{color,soul}
\usepackage{xcolor}
\usepackage{subcaption}
\usepackage{multirow}
\usepackage{booktabs}
\usepackage[moderate,leading=normal]{savetrees}

\begin{document}


\title{Training Neural Networks for Execution on Approximate Hardware}
\author{Tianmu Li}
\affiliation{%
  \institution{University of California, Los Angeles}
  \city{Los Angeles}
  \country{USA}
}
\author{Shurui Li}
\affiliation{%
  \institution{University of California, Los Angeles}
  \city{Los Angeles}
  \country{USA}
}
\author{Puneet Gupta}
\affiliation{%
  \institution{University of California, Los Angeles}
  \city{Los Angeles}
  \country{USA}
}

\begin{abstract}
Approximate computing methods have shown great potential for deep learning. Due to the reduced hardware costs, these methods are especially suitable for inference tasks on battery-operated devices that are constrained by their power budget. However, approximate computing hasn't reached its full potential due to the lack of work on training methods. In this work, we discuss training methods for approximate hardware. We demonstrate how training needs to be specialized for approximate hardware, and propose methods to speed up the training process by up to 18X. 
\end{abstract}


\maketitle

\newcommand*{\MyPath}{./}
\section{Introduction}

Machine learning using neural networks has grown in popularity in recent years. Neural networks have grown in size and complexity to improve accuracy. Due to the increasing model complexity, approximate computing methods including approximate arithmetic, stochastic computing, and analog computing have been proposed to trade some computation accuracy for better performance. However, errors introduced by approximate computing methods affect inference accuracy and make it difficult to use models trained for floating-point or fixed-point computation. Very few previous works have focused on the training techniques needed for approximate hardware. As a result, approximate computing has been limited to simple models and datasets or has to sacrifice performance to maintain accuracy.

In this work, we propose methods to improve the training of models using approximate computing. Our contributions include:
\begin{itemize}
    \item Use activation functions to approximate computation error in the backward pass.
    \item Use error injection to reduce the time per training iteration by up to 36.6X.
    \item Combine error injection with accurate modeling to maintain model accuracy.
    \item Reduce end-to-end training time by up to 18X, allowing training of models previously impossible to train on a single consumer GPU.
\end{itemize}
\section{motivation} \label{sec:motivation}

In this section, we discuss the benefits of computation with error for neural networks and motivate the need for specialized training.

\subsection{Inaccurate computation for neural networks}
While there are multiple ways to improve computation performance with inaccurate computation, we focus our study on three types:
\begin{enumerate}
    \item Stochastic computing (SC). SC represents numbers using randomized bit streams and allows single-gate additions and multiplications. The low area of stochastic computing allows high compute density and improved memory efficiency, offering up to 38.7X higher energy efficiency compared to traditional fixed-point accelerators \cite{Romaszkan2020Acoustic}.
    \item Approximate arithmetic. Approximate arithmetic focus on reducing the cost of expensive computation, typically in the form of approximate multiplication, without reducing the input and computation bit width. Removing components from an accurate multiplier reduces area and power at the cost of increased error \cite{Kulkarni2011Approx}. By introducing different levels of inaccuracy, approximate computing can achieve various accuracy and performance levels, reducing the power and area of a multiplier by 4X or more for extreme cases \cite{Mrazek2017Evo}.
    \item Analog computing. There is a growing interest in analog neural network accelerators including processing-in-memory (PIM) \cite{cimpim, drampim} and on-chip photonics \cite{shiflett2021albireo, li2022photofourier}. The main advantage of analog accelerators is that they require significantly less power to compute dot products compared to their digital counterparts, making the overall computation more power efficient, and can be more than 10X more power efficient than modern digital accelerators \cite{shiflett2021albireo}. Analog accelerators usually consist of multiple arrays, and each array computes a single dot product. Analog-to-digital converters (ADC) are required to convert the dot product results from the analog domain back to the digital domain. 
\end{enumerate}
Most of the mentioned methods can build upon traditional model compression methods like weight pruning and weight/activation quantization to further improve execution efficiency.

\subsection{Need for specialized training}
Neural networks typically assume no error during training, and cannot work reliably when errors are introduced into the computation. Stochastic computing is random by nature, and SC accumulation using bit-wise adders cannot perform accurate accumulation \cite{Romaszkan2020Acoustic}. Approximate multiplication introduces errors in multiplications by design and can have very high errors for specific input combinations \cite{Kulkarni2011Approx}. Limited by the array size, in many cases, an analog accelerator cannot compute the entire convolution in a single array, instead, partial sums will be computed \cite{cimquantization1}. The partial sum will then be quantized by the ADC before further accumulation in the digital domain since ADCs have limited precision and range. However in normal training partial sums will not be quantized, and can introduce significant errors if ADC bitwidth is low, making learned weights invalid for the actual hardware. All of these effects need to be taken care of during training. As shown in Tab. \ref{tab:ae_model_acc}, running a model pretrained for fixed-point computation directly with approximate computing can drop accuracy by 8-57\% pts compared to modeling the computation properly during training.

Modeling approximate computing accurately in the forward pass is expensive compared to accurate multiplication and addition, as shown in Tab. \ref{tab:ae_op_cost}. In practice, the runtime difference can be even larger, since floating-point conv2d/linear layers can use optimized kernels from libraries like cuDNN \cite{nvidia2014cudnn}, while functions to model approximate computing methods require additional coding (i.e., implementing custom kernels). As a result, \textbf{specialized methods to reduce training time are necessary for approximate hardware.}

\begin{table}[hbp]
    \centering
    \caption{Relative multiplication and addition cost. FP32 multiplication and addition are used as the baseline. The number of operations in C++ is used as the cost value.}
    \resizebox{\columnwidth}{!}{
    \begin{tabular}{l|c|c}
        \toprule
        Method & Multiplication & Addition \\
        \hline
        Floating point & 0.5(fused) & 0.5(fused)\\
        \hline
        \multirow{2}{*}{Stochastic Computing (32-bit)} & 64(unrolled) & 64(unrolled) \\
        & 2(packed) & 2(packed) \\
        \hline
        Approximate Multiplication & 86 & 1 \\
        \hline
        \multirow{2}{*}{Analog Computing} & \multirow{2}{*}{1} & 1(within channel) \\
         & & 9(between channel) \\
        \bottomrule
    \end{tabular}
    }
    \label{tab:ae_op_cost}
\end{table}

\subsection{Previous works on training for approximate hardware.}
Due to the difficulties in modeling computation errors during training, inaccurate computing methods have not been able to fully utilize their benefits in previous works. For analog computing, there are a few works discussing the quantization of partial sums (or ADCs) of PIMs \cite{cimquantization1, cimquantization2} and photonic accelerators \cite{photonicquantization1, photonicquantization2}. However, the main objective of these works is accuracy optimization using their proposed quantization methods and they usually focus on simpler datasets and models so that runtime is relatively manageable. The runtime issue is not addressed in these works, which is the primary focus of our work.  

Previous works have also tried to model the computation error accurately during training, but the expensive emulation cost prevents training of more complicated models \cite{Li2021Geo} in the case of stochastic computing, or limits the approach to computation suitable for mapping to commercial training hardware (typically GPUs) \cite{Jing2022Approx} in the case of approximate multiplication. Other works try to reduce the error of approximation so that models trained for fixed-point computation can be used directly for inference. For stochastic computing, SC additions are replaced with fixed-point additions to reduce error \cite{Sim2017BISC}. For approximate multiplication, approximation is limited to locations that don't affect model accuracy too much \cite{Zhang2015Approx, Venkataramani2014Axnn}. Since approximate computing relies on the error for performance, reducing the error comes at the cost of diminished performance benefits.
\section{methodology} \label{sec:methodology}

Our training improvements include three components: activation approximation for non-linear computation, error injection during training with fine-tuning, and gradient checkpointing. We will discuss the three components in detail in this section. We will use the smaller CIFAR-10 dataset for most of this section due to the runtime limitations on larger datasets. This setup limits the performance benefits of our methods and also makes the runtime results less reliable due to large run-to-run variations, but we will demonstrate the full performance benefits in Sec. \ref{sec:results}. We focus on three types of approximate computing methods mentioned in Sec. \ref{sec:motivation} to showcase the benefits of our approaches:
\begin{itemize}
    \item Stochastic computing. We use linear feedback shift registers for stream generation, AND gate for multiplication and OR for addition, which is similar to the setup in \cite{Romaszkan2020Acoustic}. We use 32-bit split-unipolar streams (64 total bits).
    \item Approximate multiplier. We use an approximate multiplier from EvoApproxLib \cite{Mrazek2017Evo}. Specifically, we use the mul7u\_09Y setup. It is a 7-bit unsigned multiplier in the paretal-optimal set for mean-relative error, which is suitable for neural networks and has 8-bit input precision when combining the sign bit. Compared to an accurate 7-bit multiplier, the approximate multiplier has ~4X lower error and power consumption. 
    \item Analog computing. To make a more generic argument, we set the effective array size of analog accelerators such that the partial sum of every convolution channel is quantized by the ADC. That is 9 for Resnet-tiny (Resnet-18 shrunk for TinyML applications used in \cite{mlperftiny}) and Resnet-18 \cite{resnet}, and 25 for TinyConv (a four-layer CNN used in \cite{Lai2018MlCmsis}). Since analog accelerators usually only support positive inputs and weights, we use split-unipolar accumulation in our training setup to handle negative weights, which results in $2\times$ computation. We assume 4-bit ADCs are used for all evaluations in this paper. A relatively low ADC bitwidth is selected since the ADCs are usually the power bottlenecks of analog accelerators, therefore low-biwidth ADCs are always preferred if the accuracy impact is not significant. The bitwidth for inputs and weights is set to 8-bit for all cases to focus on the impact of partial sum quantization.
\end{itemize}
While there are other approximate hardware implementations, we limit our study to the above three methods as they cover a wide variety of approximate computing. Our method is applicable to other approximate hardware setups with minimal change.

\subsection{Approximation Proxy Activation} \label{sec:meth_activation}
One issue of inaccurate computation is the non-linearity introduced to multiply-accumulate operations, which is separate from non-linear activation functions like ReLU. We use stochastic computing and analog computing as examples. Approximate multiplier does not suffer from the non-linearity issue, as error is only introduced during multiplication. For SC, if the two input values $a$, and $b$ are uncorrelated, the OR adder performs $a+b-ab$ on average. For analog computing, partial sum and output results need to be clamped and quantized. Accurately modeling the imperfections in computation can be costly in the backward pass. For instance, the previously-mentioned OR adder requires tracking almost all inputs in the adder ($\frac{\partial}{\partial a_i} \text{OR}(a_j)=\prod_{j\neq i}(1-a_j)$) during backpropagation, whereas accurate addition is much simpler in the backward pass (partial derivative=1). On top of being expensive to model, the nonlinearity can hinder convergence if it is not taken into consideration. Using activations as a proxy during backward pass is first proposed in\cite{Romaszkan2020Acoustic} to simulate the effect of SC OR accumulation. \textit{Contrary to normal activation functions, the added activation proxy is only used during training in the backward pass and is not used during inference.} With proper implementation of the activation function, the overhead of the activation function can be negligible. As is shown in Tab. \ref{tab:acc_act}, modeling the non-linearity using an activation function is necessary. TinyConv is a four-layer CNN used in \cite{Lai2018MlCmsis}, and Resnet-tiny is Resnet-18 shrunk for TinyML application used in \cite{mlperftiny}. Models are trained with accurate modeling of stochastic and analog computing in the forward pass. For analog computing, accuracy is noticeably lower when trained without the activation function. For stochastic computing, training does not converge at all without the activation function. \par


\begin{table}[]
    \centering
    \caption{Accuracy benefits of using activation functions.}
    \begin{tabular}{l|c|c}
        \toprule
        Setup & TinyConv & Resnet-tiny \\
        \hline
        \multicolumn{3}{c}{Stochastic Computing} \\
        \hline
        No Activation & 41.64\% & 10.00\%\\
        With Activation & 72.18\% & 79.76\% \\
        \hline
        \multicolumn{3}{c}{Analog Computing (4-bit)} \\
        \hline
        No Activation & 74.85\% & 69.21\% \\
        With Activation & 79.20\% & 77.23\% \\
        \bottomrule
    \end{tabular}
    \label{tab:acc_act}
\end{table}



\begin{figure}
    \centering
    \begin{subfigure}[b]{0.48\columnwidth}
        \includegraphics[width=\textwidth]{\MyPath/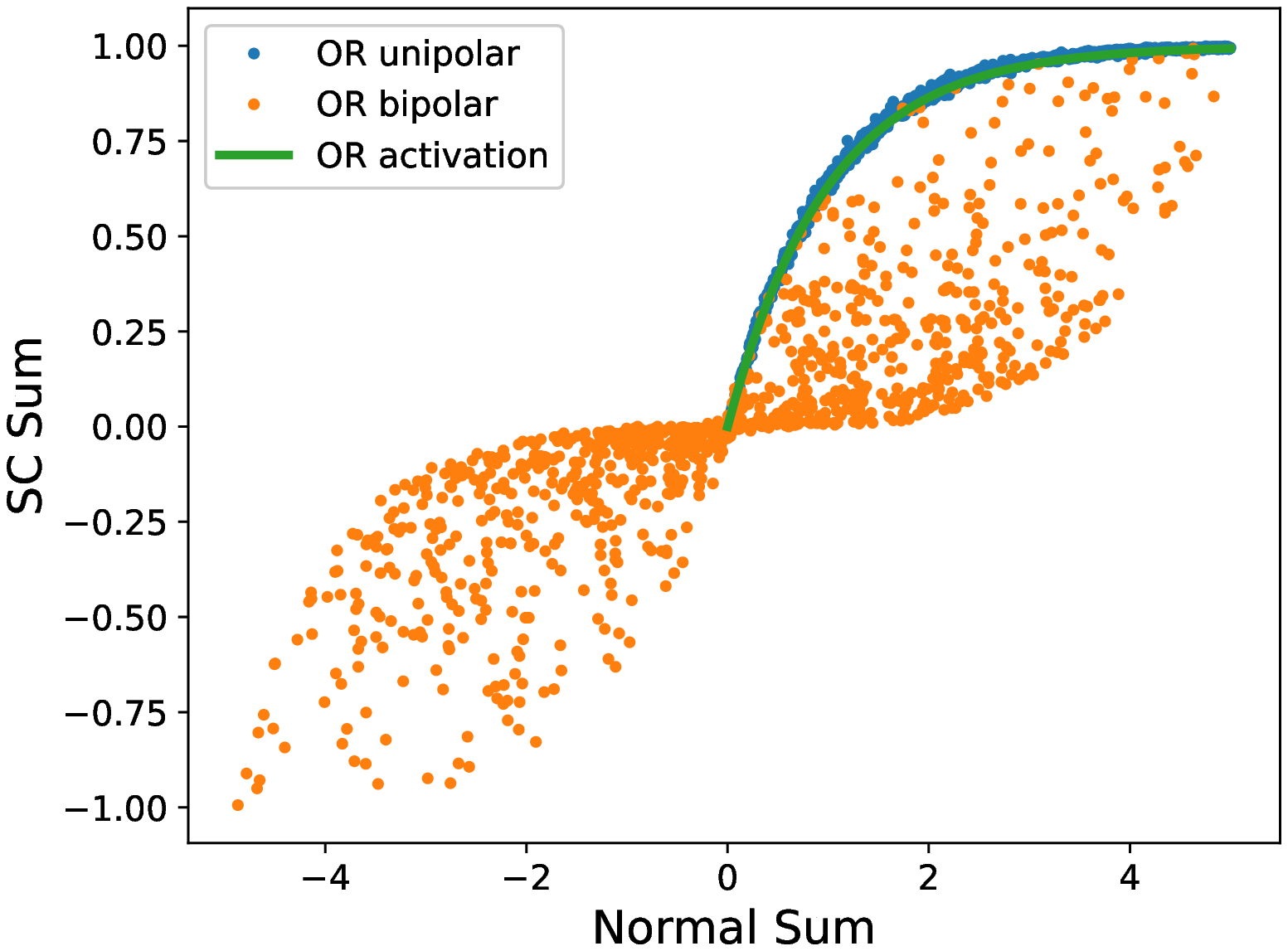}
        \caption{}
        \label{fig:or_uni_bi}
    \end{subfigure}
    \hfill
    \begin{subfigure}[b]{0.48\columnwidth}
        \includegraphics[width=\textwidth]{\MyPath/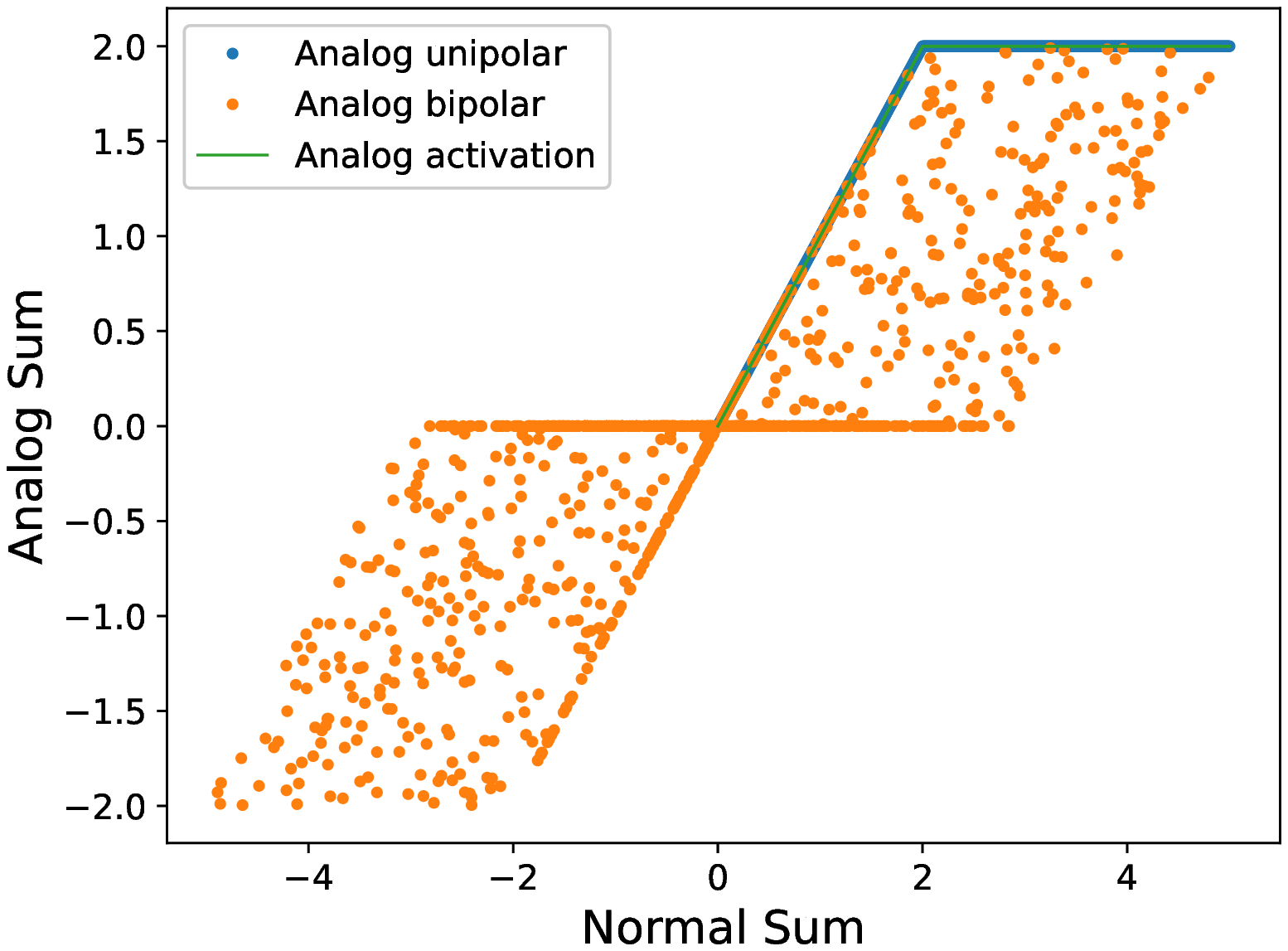}
        \caption{}
        \label{fig:cim_uni_bi}
    \end{subfigure}
    \caption{Activation modeling behavior of unipolar and bipolar (a) stochastic and (b) analog computation. The bipolar versions performs subtraction between two unipolar (positive and negative) inputs to achieve the full range. For analog computation, the ADC saturation is modeled as a clamp at 2 in this example, and other effects (e.g. size of accumulation) are not considered.}
    \label{fig:comb_uni_bi}
\end{figure}
Using an activation function requires the computation to be (mostly) associative. Take the previously used OR accumulation and analog computing as examples. Computation is broken up into positive and negative parts since both work on unipolar (positive-only) inputs. Accumulation within each part is associative, but subtracting the two parts isn't. This behavior can be visualized in Fig. \ref{fig:comb_uni_bi}. While the activation function can approximate unipolar OR accumulation, a single activation cannot be used to model the entire accumulation when subtraction is factored in. As such, accumulations need to be split into positive and negative parts in the backward pass, so that each part can be modeled using accurate accumulation with an activation function. Similar effects can be seen in the analog computing setup. Since the positive and negative parts saturate individually, a single activation function is also insufficient. Tab. \ref{tab:meth_act_both} lists the activation functions used for stochastic computing and analog computing.

\begin{table}[htp]
    \centering
    \caption{Activation functions for stochastic computing and analog computing. $x$ is the output of a layer before activation. $x_{pos}$ is the output of positive weights and $x_{neg}$ is the output of negative weights. Inputs are assumed to be non-negative due to ReLU activation.}
    \resizebox{\columnwidth}{!}{
    \begin{tabular}{l|c}
        \toprule
        Method & Activation Function \\
        \hline
        Stochastic Computing &  $\textnormal{SC\_act}(x) = (1-e^{-x_{pos}}) - (1-e^{-x_{neg}})$\\
        Analog Computing & $\textnormal{Analog\_act}(x) = \textnormal{HardTanh}(x_{pos}) - \textnormal{HardTanh}(x_{neg})$\\
        \bottomrule
    \end{tabular}
    }
    \label{tab:meth_act_both}
\end{table}

\subsection{Error injection} \label{sec:meth_ae}
Despite resolving the issue of backpropagation, the activation function method cannot fully replace the forward propagation. Training models for non-floating-point computation typically involves modeling the computation accurately in the forward pass, be it low-precision fixed-point computation (including extreme precision like binarization \cite{Rastegari2016XNOR}) or approximate computation \cite{Jing2022Approx}. For fixed-point computation, modeling the computation in the forward pass is relatively cheap. An element-wise fake-quantization operator followed by a normal convolution/linear operator is sufficient. The same cannot be said for approximate computing methods. SC requires emulating the stream generation and bit-wise multiplication in hardware. The approximate multiplier described in Sec. \ref{sec:motivation} is made up of hundreds of lines of bit manipulation. Analog computing requires emulating the limited hardware array size and ADC precision during computation. Tab. \ref{tab:ae_op_cost} demonstrates the high cost of emulating approximate computing. In all cases, emulating the computation is expensive, requires additional coding, and cannot be completely overcome even with abundant programming resource.

\begin{table}[hbp]
    \centering
    \caption{Accuracy impact of modeling approximate computation. "With Model" models the approximate computation method accurately in the forward pass.}
    \resizebox{\columnwidth}{!}{
    \begin{tabular}{l|c|c|c|c}
        \toprule
         & \multicolumn{2}{c|}{TinyConv} & \multicolumn{2}{c}{Resnet-tiny} \\
        Method & Inference Only & With Model & Inference Only & With Model \\
        \hline
        Stochastic Computing & 14.87\% & 72.18\% & 48.57\% & 79.76\% \\
        Approximate Multiplication & 71.88\% & 83.35\% & 76.95\% & 85.25\% \\
        Analog Computing (4b) &  58.43\% & 78.94\% & 47.15\%  & 77.23\%    \\
        \bottomrule
    \end{tabular}
    }
    \label{tab:ae_model_acc}
\end{table}

While it is expensive to model approximate computing accurately in the forward pass, skipping modeling degrades accuracy, as shown in Tab. \ref{tab:ae_model_acc}. Despite being sufficient for the backward pass, the activation method defined in Sec. \ref{sec:meth_activation} is not sufficient for the forward pass. To combat this limitation, we propose to replace accurate modeling with two types of error injection coupled with normal training. \par
\begin{figure}
    \centering
    \begin{subfigure}[b]{0.45\columnwidth}
    \centering
        \includegraphics[width=\textwidth]{\MyPath/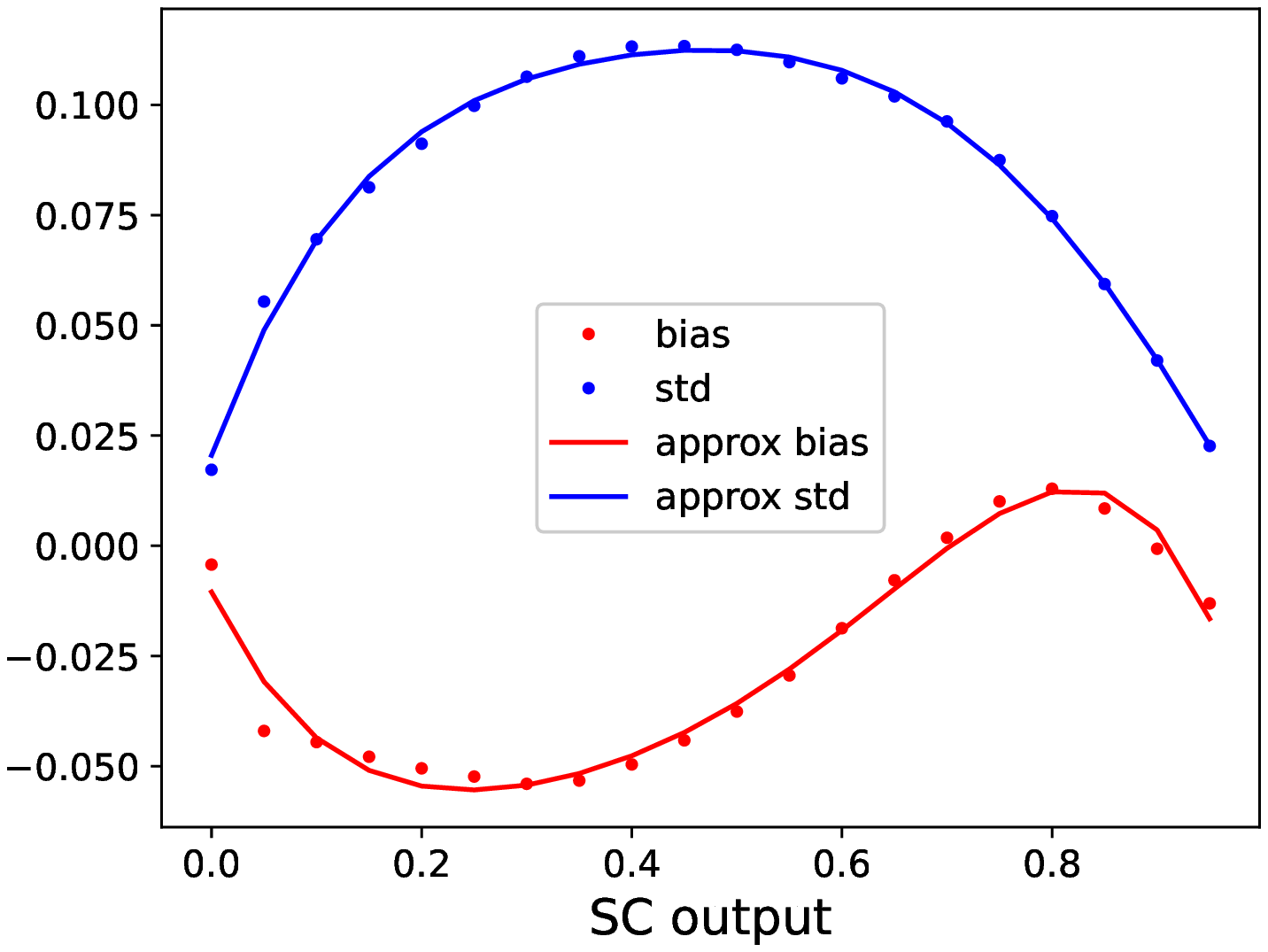}
        \caption{Layer 1}
        \label{fig:err_l1}
    \end{subfigure}
    \hfill
    \begin{subfigure}[b]{0.45\columnwidth}
        \includegraphics[width=\textwidth]{\MyPath/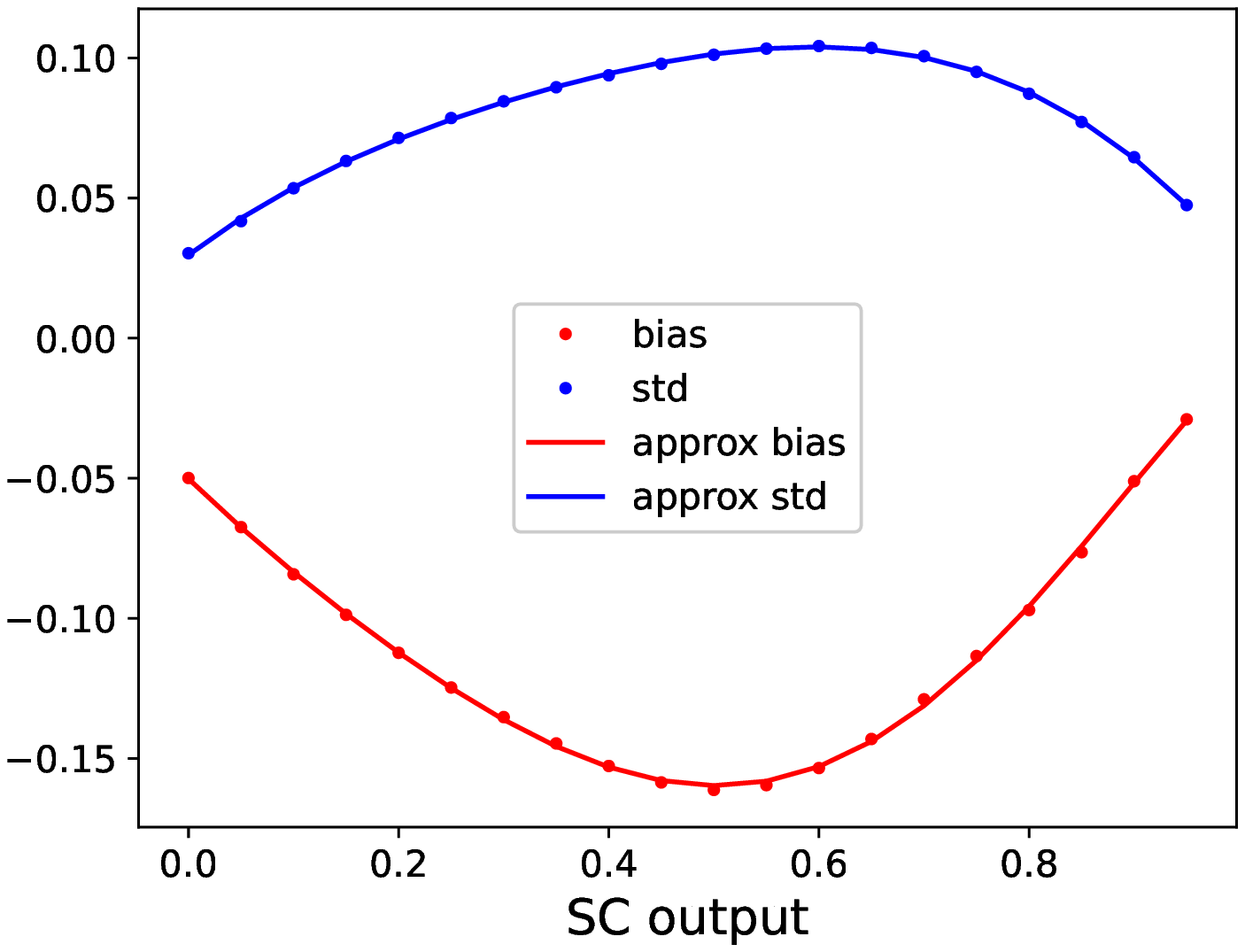}
        \caption{Layer 2}
        \label{fig:err_l2}
    \end{subfigure}
    \hfill
    \begin{subfigure}[b]{0.45\columnwidth}
        \includegraphics[width=\textwidth]{\MyPath/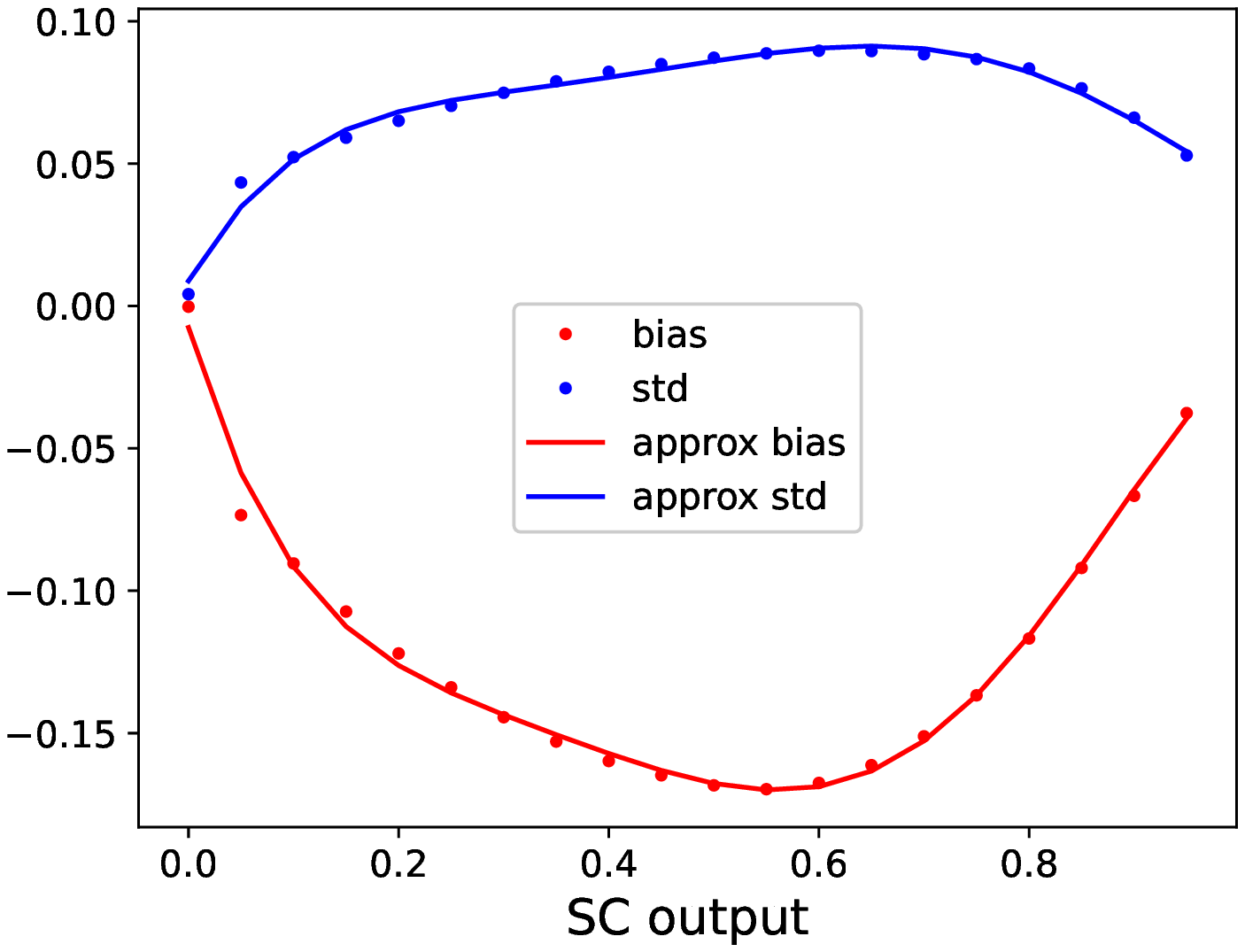}
        \caption{Layer 3}
        \label{fig:err_l3}
    \end{subfigure}
    \hfill
    \begin{subfigure}[b]{0.45\columnwidth}
        \includegraphics[width=\textwidth]{\MyPath/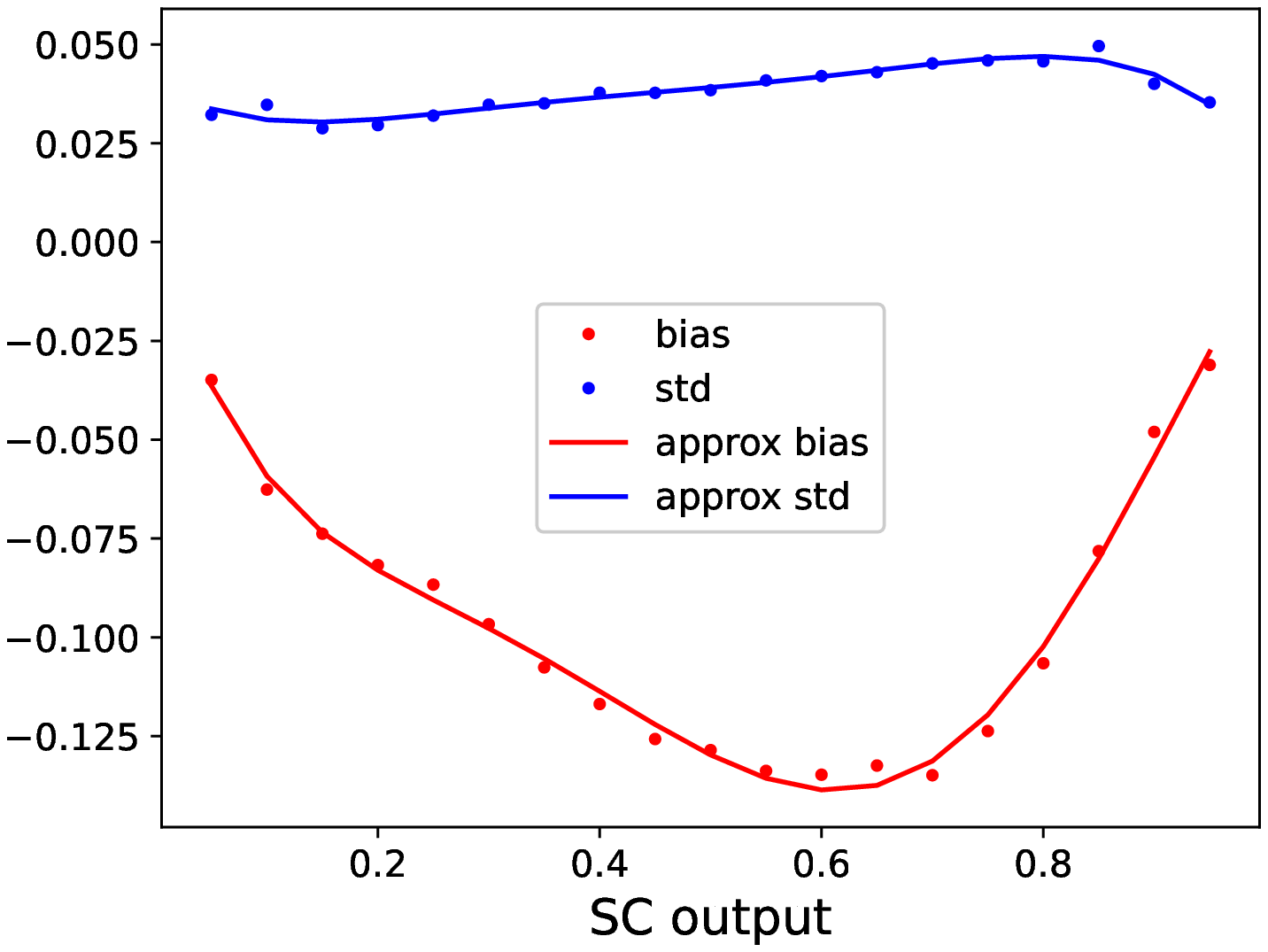}
        \caption{Layer 4}
        \label{fig:err_l4}
    \end{subfigure}
    \caption{Difference between outputs from stream computation and normal computation+activation.}
    \label{fig:err_layer}
\end{figure}

\paragraph{Type 1}
The first type of error injection models the difference between activation function and accurate modeling as functions of the output values of a layer, and we use it for stochastic computing and approximate multiplication. Fig. \ref{fig:err_layer} shows the error average and variation of the four layers of TinyConv using stochastic computing. Compared to the value expected from the simple activation function $y=1-e^{-x}$, the actual layer output differs significantly. The variance (represented as "std") in the plots means that it is impossible to fully capture the computation details using only an activation function. Given that the activation function is only an approximation by design, it cannot capture all the characteristics of the inaccurate computation. On the other hand, the non-zero average error means that the activation function doesn't represent the target computation accurately on average. Activation functions are derived under certain assumptions. In the case of OR accumulation, the assumption is that the size of accumulation $n$ is large and all input values are small and similar in value. These assumptions don't always hold in a real model, and the difference in average error between layers means it's impossible to have a single activation function for the entire model. For deep models, the error of the activation function accumulates and results in non-usable models after training. To resolve this issue, we propose to add correction terms to the activation function during training.

Take the error profile shown in Fig. \ref{fig:err_layer} as an example. The mean and variance of the error are plotted with respect to the output after the activation function, and both appear to be smooth functions. From this observation, we model the average as a function of activated output values on top of the original activation function, and the variance as a normally distributed random error with variance dependent on the activated value. Both curves are fitted to a polynomial function that's different for each layer and calibrated 5 times per epoch. Though it is possible to lower it further, this frequency sufficiently amortizes the cost of calibration.

\paragraph{Type 2}
The second type of error injection is used in training analog accelerators. The method is similar to the mentioned error injection method used for stochastic computing and approximate multiplication, but with slight differences. In this case, the overall quantization error of output activations (the sum of individual partial sum quantization errors) is modeled and injected. We first perform an accurate forward pass for one batch, which quantizes the partial sum of every channel before accumulation. The accurate results are subtracted from the results separately generated by the normal Conv2d function (does not quantize partial sums) to obtain the actual overall quantization error. Then the mean and variance of the quantization error are calculated and stored for each convolution layer. Instead of obtaining the statistics on a per activation or per filter granularity, we only calculate a single mean and variance for an entire layer. There are two reasons for this choice: 1) empirical results suggest that the accuracy of a single mean and variance is better than calculating mean and variance at other granularity, and 2) memory saving, since only two values need to be stored per layer. We calibrate the error statistics every 10 batches, which achieves a good trade-off between accuracy and runtime. 
For the majority of batches that do not require calibration, we only compute a normal Conv2d in the forward pass (partial sums not quantized). Then we simulate the quantization error by generating random error according to a normal distribution with mean and variance set to the values obtained from the last calibration batch. The simulated quantization error is directly added to the Conv2d outputs. This way, for non-calibration batches, the runtime should be almost identical to normal convolution, since generating quantization error takes significantly less time compared to the convolution operation.

Using error injection in the place of accurate modeling improves accuracy compared to not modeling at all, as shown in Tab. \ref{tab:ae_error_acc}. However, training with error injection alone is not sufficient to completely bridge the gap, especially for analog computing with a low ADC bitwidth. Later we will show the accuracy gap can be eliminated through a fine-tuning step. Runtime is improved up to 6.7X with error injection compared to using an accurate model during training, as shown in Tab. \ref{tab:ae_error_time}. In most cases, the error injection runtimes are comparable to those in normal training (the "Without Model" column). The slightly longer error injection runtime for analog computing is due to its higher calibration frequency. On relatively a larger network and dataset (Resnet-18 on Imagenet), the runtime improvement of error injection can be larger for analog computing (Shown in Sec. \ref{sec:results}).  \par

\begin{table*}[htp]
    \centering
    \caption{Accuracy impact of error injection training.}
    \resizebox{\textwidth}{!}{
    \begin{tabular}{l|c|c|c|c|c|c|c|c}
        \toprule
        & \multicolumn{4}{|c}{TinyConv} & \multicolumn{4}{|c}{Resnet-tiny} \\
        \hline
        Method & Inference Only & With Model & Error Injection & Fine-tuning & Inference Only & With Model & Error Injection & Fine-tuning \\
        \hline
        Stochastic Computing & 14.87\% & 72.18\% & 64.01\% & 73.09\% & 48.57\% & 79.76\% & 76.71\% & 81.29\% \\
        Approximate Multiplication & 71.88\% & 83.35\% & 79.06\% & 83.05\% & 76.95\% & 85.25\% & 81.06\% & 84.85\% \\
        Analog Computing (4b) &  58.43\% &  78.94\% & 62.02\% & 77.95\% &47.15\%  &  77.23\% & 71.10\%  & 76.05\%\\
        \bottomrule
    \end{tabular}
    }
    \label{tab:ae_error_acc}
\end{table*}

\subsection{Fine tuning} \label{sec:meth_tune}
\begin{figure*}
    \centering
    \begin{subfigure}[b]{0.33\textwidth}
        \centering
        \includegraphics[width=\textwidth]{\MyPath/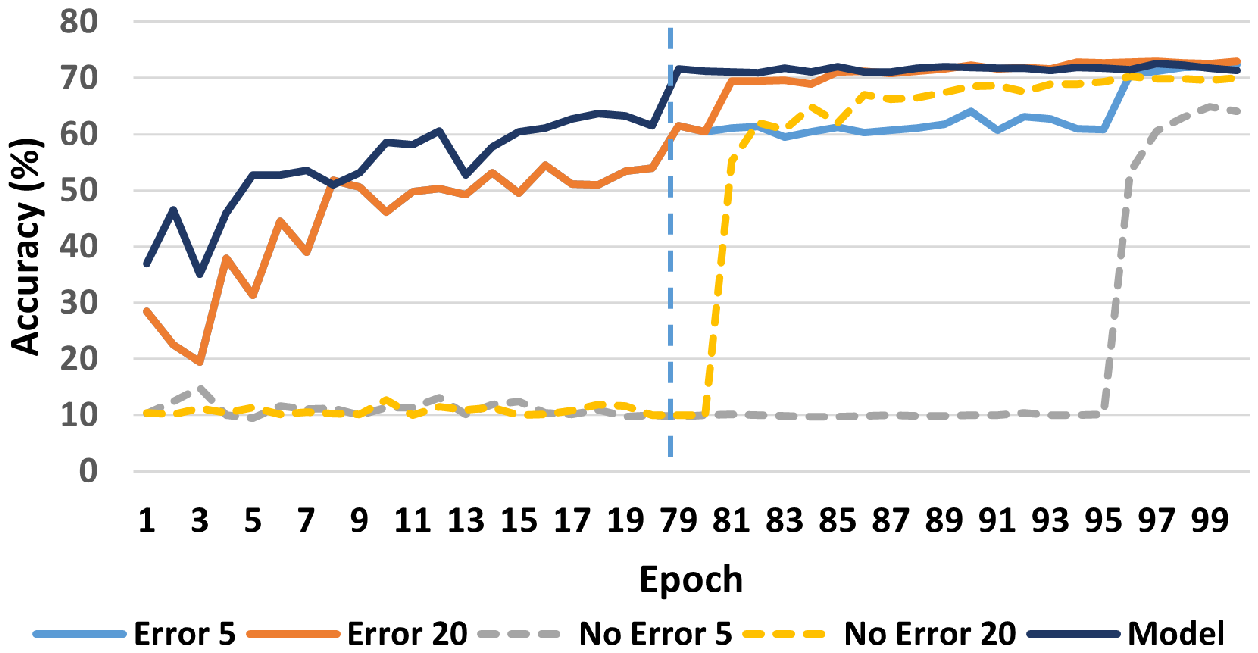}
        \caption{}
        \label{fig:conv_or1_tinyconv}
    \end{subfigure}
    \hfill
    \begin{subfigure}[b]{0.33\textwidth}
        \centering
        \includegraphics[width=\columnwidth]{\MyPath/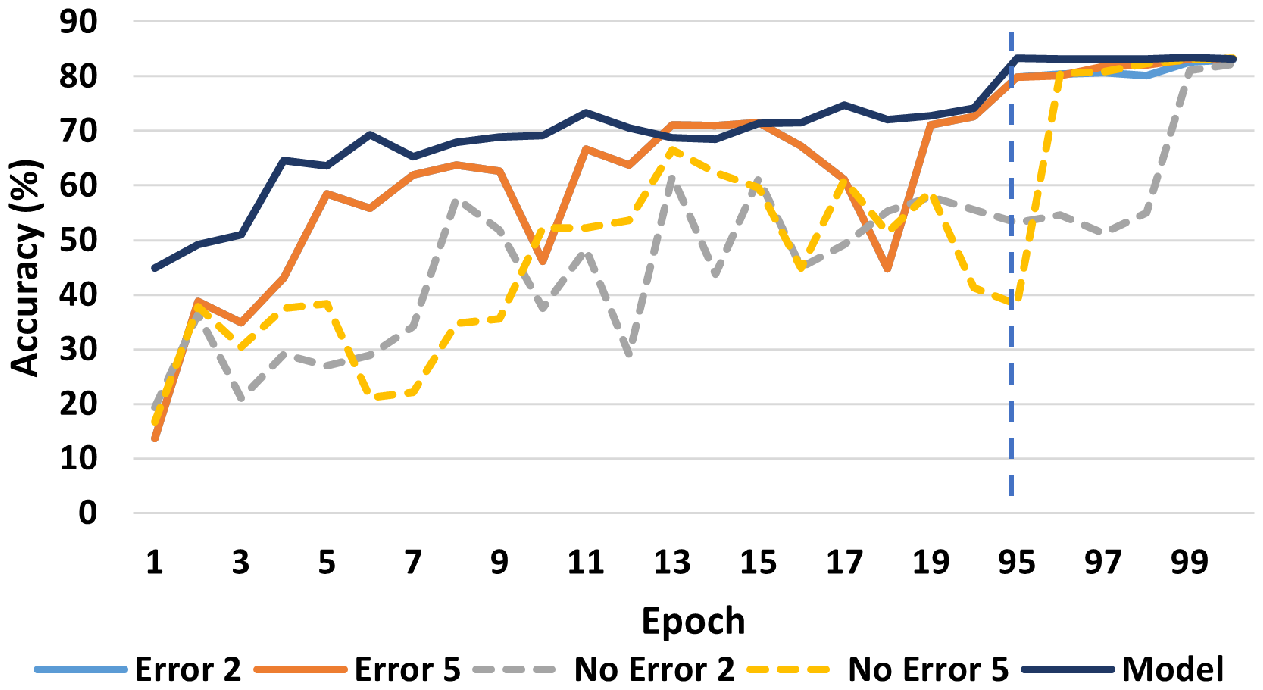}
        \caption{}
        \label{fig:conv_evo_tinyconv}
    \end{subfigure}
    \hfill
    \begin{subfigure}[b]{0.3\textwidth}
        \centering
        \includegraphics[width=\columnwidth]{\MyPath/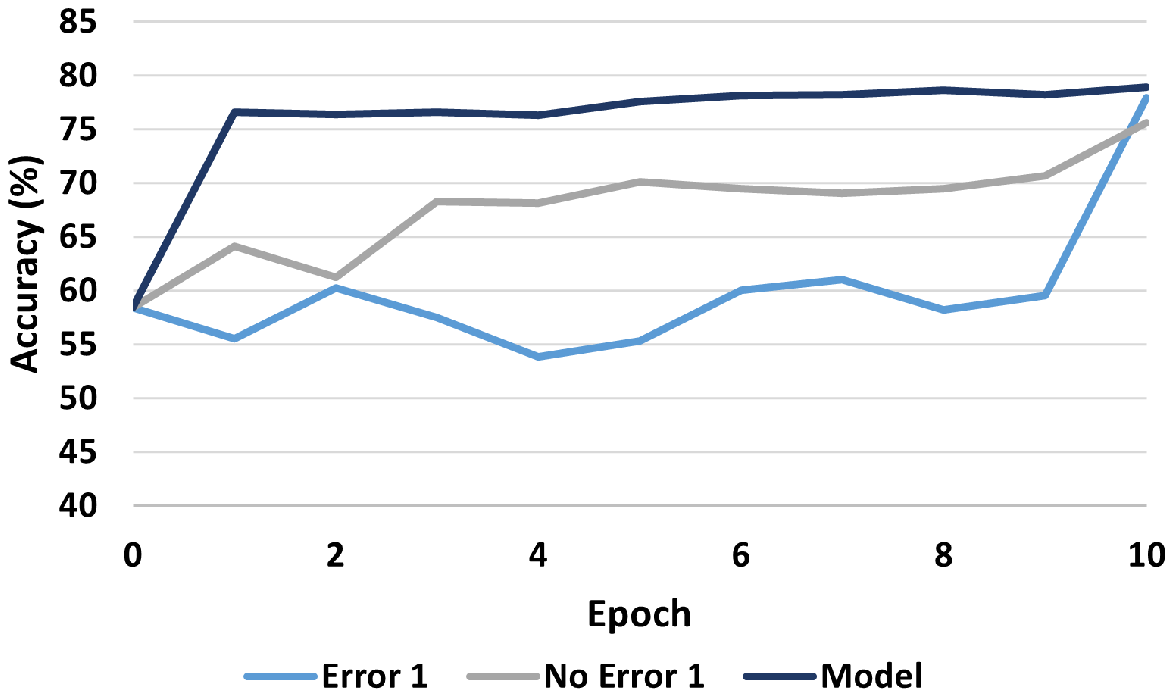}
        \caption{}
        \label{fig:conv_analog_tinyconv}
    \end{subfigure}
    \caption{Convergence behavior of TinyConv with and without error injection using (a) stochastic computing, (b) approximate multiplication, and (c) analog computing. Stochastic computing and approximate multiplication trains from scratch for 100 epochs and the first 20 epochs and the last few epochs are shown. Analog computing trains from an INT-8 pretrained weight for 10 epochs since training from scratch is not required.  "Error X" means training with error injection with X epochs of fine-tuning with an accurate model". "No Error X" means training without error injection with X epochs of fine-tuning. "Model" means accurate modeling throughout training.}
\end{figure*}

While the error injection mentioned in Sec. \ref{sec:meth_ae} cannot achieve the same accuracy as training with accurate modeling, it reduces the amount of fine-tuning with accurate modeling required to achieve the same accuracy. The goal of error injection is to simulate the error during training, which can make the model more robust to errors. However, since the injected error is randomly generated while the actual error is input-dependent, error injection cannot achieve the same accuracy as accurate modeling. Still, error injection makes the model more robust such that with a small amount of accurate modeling, the model weights can quickly converge to the optimal point.
Fig. \ref{fig:conv_or1_tinyconv} compares the convergence behavior when error injection is combined with accurate modeling for SC when trained for CIFAR-10. For stochastic computing, error injection combined with 5 epochs of fine-tuning is sufficient to achieve the same accuracy as using an accurate model throughout the training. In contrast, convergence is poor without error injection, and the model does not converge to the same accuracy even with 20 epochs of fine-tuning.

Fig. \ref{fig:conv_evo_tinyconv} compares the convergence behavior for approximate multiplier on the same model. Since the accuracy gap is smaller with approximate multiplier, training without error injection can also converge properly with 5 epochs of fine-tuning. However, error injection reduces that further to 2 epochs.

In the case of analog computing with 4-bit partial sum quantization, training with an accurate model usually converges between 10-15 epochs for the CIFAR-10 dataset from pretrained INT8 weights. It is not necessary to train from scratch using accurate model or error injection.  Empirical results of analog computing training suggest that using one-fourth of an epoch can achieve the same accuracy compared to fine-tuning with a complete epoch. Therefore only the last one-fourth epoch is used for fine-tuning, which adds minimum runtime overheads. Unlike SC or approximate multiplication, fine-tuning cannot fully recover the accuracy gap for a low ADC bitwidth, but the accuracy drop compared to the accurate model is usually small (around 1\%). Fig. \ref{fig:conv_analog_tinyconv} compares the convergence behavior for analog computing, with calibration and fine-tuning, error injection can achieve similar accuracy compared to accurate model. Without error injection, even keeping the same calibration and fine-tuning setup as the error injection version, the training fails to converge to the same accuracy as the version with error injection. 

Overall, fine-tuning on top of error injection removes the accuracy deficit of error injection alone. As shown in Tab. \ref{tab:ae_error_acc}, accuracy after fine-tuning ("Fine-tuning" columns) is at most 1\% pt lower than training completely using an accurate model.


\subsection{Gradient checkpointing} \label{sec:checkpoint}
Both the activation function in Sec. \ref{sec:meth_activation} and error injection in Sec. \ref{sec:meth_ae} add additional steps in the computational graph during training, which increases memory requirement and limits batch size for larger models. To reduce memory consumption, we use the gradient checkpointing setup from \cite{Chen2016Gradient}. Since the added functions are point-wise functions and have low compute intensity ($<20$ OPS/memory access), checkpointing all added computations have minimal effect on runtime even when not memory bound. Tab. \ref{tab:meth_ckpt} compares the runtime and memory consumption with and without checkpointing for Resnet-18 on the ImageNet dataset. Checkpointing allows training with a larger batch size, which improves GPU efficiency and reduces runtime required per epoch by 22\% in this case. 

\begin{table}[hbp]
    \centering
    \caption{Training runtime and memory requirement of stochastic computing without and without gradient checkpointing. We use the maximum batch size achievable which is a power of 2.}
    \resizebox{\columnwidth}{!}{
    \begin{tabular}{l|c|c|c}
        \toprule
        Setup & Memory (MB) & Batch Size & Runtime (s/epoch) \\
        \hline
        With Checkpoint & 19840 & 256 & 1326 \\
        Without Checkpoint & 12766 & 128 & 1692 \\
        \bottomrule
    \end{tabular}
    }
    \label{tab:meth_ckpt}
\end{table}
\section{results} \label{sec:results}

In this section, we demonstrate the benefits of our techniques on a more complicated workload. We use Resnet-18 training on the ImageNet dataset to demonstrate the benefits. Models are trained on a single RTX 3090 using mixed precision. We use PyTorch 1.12 as the baseline framework, and implement the additional operators as CUDA C++ extensions, including accurate modeling for stochastic computing, approximate multiplication, and analog computing. While we make the best effort to optimize the kernels used for modeling approximate computing, we cannot guarantee that the implementations are optimal. Models are trained from a checkpoint provided by PyTorch to reduce training time, which uses fp32. Because of this, the number of epochs required to converge is different between the three approximate computing setups. Tab. \ref{tab:training_setup} lists the detailed training setup.

\begin{table}[hbp]
    \centering
    \caption{Runtime impact of error injection training. Shown is the time (seconds) required per epoch. Runtimes are measured on an RTX 3090 using TF32 precision and batch size=256. SC and approximate multiplication are slower due to the need to split positive and negative computations discussed in Sec. \ref{sec:meth_activation}. Fine-tuning runtime is the same as the runtime with accurate model (the "With Model" column).}
    \resizebox{\columnwidth}{!}{
    \begin{tabular}{l|c|c|c}
        \toprule
        Method & Without Model & With Model & Error Injection \\
        \hline
        & \multicolumn{3}{|c}{TinyConv} \\
        \hline
        Stochastic Computing & 3.86 & 9.50 & 3.90\\
        Approximate Multiplication & 3.86 & 28.3 & 4.20\\
        Analog Computing (4b) & 2.13 & 3.91  & 2.73 \\
        \hline
        & \multicolumn{3}{|c}{Resnet-tiny} \\
        \hline
        Stochastic Computing & 8.51 & 13.3 & 8.65\\
        Approximate Multiplication & 8.13 & 38.4 & 9.11\\
        Analog Computing (4b) & 4.88 & 8.51 & 6.53\\
        \bottomrule
    \end{tabular}
    }
    \label{tab:ae_error_time}
\end{table}

\begin{table}[hbp]
    \centering
    \caption{Epochs used for training. Methods with higher accuracy require fewer epochs of training.}
    \resizebox{\columnwidth}{!}{
    \begin{tabular}{l|c|c}
        \toprule
        Method & Error Injection & Fine-tuning \\
        \hline
        Stochastic Computing & 30 & 5 \\
        Approximate Multiplication & 34 & 1 \\
        Analog Computing (4b) & 14 & 1 \\
        \bottomrule
    \end{tabular}
    }
    \label{tab:training_setup}
\end{table}

Tab. \ref{tab:acc_imagenet} shows the accuracy achieved for all three setups. While there are no previous accuracy results that we can compare against, the accuracy results follow the same trend as that seen for smaller models discussed in Sec. \ref{sec:methodology}. Approximate multiplication and analog computing take too long to train without our improvements, as we will show later. \par

\begin{table}[htp]
    \centering
    \caption{Top-1 accuracy of Resnet-18 on ImageNet.}
    \resizebox{\columnwidth}{!}{
    \begin{tabular}{l|c|c}
        \toprule
        Method & Without Improvements & With Improvements \\
        \hline
        Stochastic Computing  & 52.45\% & 54.29\% \\
        Approximate Multiplication & N/A & 67.81\% \\
        Analog Computing (4b) & N/A & 63.40\%\\
        \bottomrule
    \end{tabular}
    }
    \label{tab:acc_imagenet}
\end{table}

Tab. \ref{tab:training_time} showcases the performance benefits of the proposed methods when combined. 
The improvements shown here underestimate the benefits of the proposed methods due to the following factors:
\begin{itemize}
    \item The "Without Improvements" column assumes that the activation function mentioned in Sec. \ref{sec:meth_activation} is used in the backward pass. If the activation function is not used, the backward pass also needs to be modeled accurately.
    \item Models are validated after each epoch. Since validation uses accurate modeling and is the same with and without improvements, the performance gap will be even more significant if the validation frequency is reduced.
\end{itemize}
Despite these limitations, our methods reduce end-to-end training time by 2.4X to 18.2X. For stochastic computing and approximate multiplication, the performance benefit roughly follows the runtime difference shown in Tab. \ref{tab:ae_error_time}, as runtime was bottlenecked by accurate modeling even for smaller models. Approximate multiplication especially benefits from error injection, as iteration time reduces by 36.6X compared to accurate modeling. For analog computing, the runtime difference on ImageNet is larger than that on CIFAR-10, as both TinyConv and Resnet-tiny cannot fully utilize the GPU on CIFAR-10. 

\begin{table}[htp]
    \centering
    \caption{End-to-end runtime improvements. Time is measured in hours required to converge. The "Without Improvements" results are estimated for models that are impossible to train without the improvements.}
    \resizebox{\columnwidth}{!}{
    \begin{tabular}{l|c|c}
        \toprule
        Method & Without Improvements & With Improvements \\
        \hline
        Stochastic Computing & 42.8 & 17.6 \\
        Approximate Multiplication & 592 & 32.4 \\
        Analog Computing (4b) & 55.3 & 12.1 \\
        \bottomrule
    \end{tabular}
    }
    \label{tab:training_time}
\end{table}

\section{conclusion}

In this work, we propose several methods to improve training performance for inference on approximate hardware. Through the use of activation modeling, error injection with fine-tuning, and gradient checkpointing, we achieve convergence on a wide range of approximate hardware. Our method makes it feasible to train large models like Resnet-18 on the ImageNet dataset using a single consumer GPU.

\bibliographystyle{ACM-Reference-Format}
\bibliography{acmart}


\begin{thebibliography}{22}


\ifx \showCODEN    \undefined \def \showCODEN     #1{\unskip}     \fi
\ifx \showDOI      \undefined \def \showDOI       #1{#1}\fi
\ifx \showISBNx    \undefined \def \showISBNx     #1{\unskip}     \fi
\ifx \showISBNxiii \undefined \def \showISBNxiii  #1{\unskip}     \fi
\ifx \showISSN     \undefined \def \showISSN      #1{\unskip}     \fi
\ifx \showLCCN     \undefined \def \showLCCN      #1{\unskip}     \fi
\ifx \shownote     \undefined \def \shownote      #1{#1}          \fi
\ifx \showarticletitle \undefined \def \showarticletitle #1{#1}   \fi
\ifx \showURL      \undefined \def \showURL       {\relax}        \fi
\providecommand\bibfield[2]{#2}
\providecommand\bibinfo[2]{#2}
\providecommand\natexlab[1]{#1}
\providecommand\showeprint[2][]{arXiv:#2}

\bibitem[Angizi et~al\mbox{.}(2018)]%
        {cimpim}
\bibfield{author}{\bibinfo{person}{Shaahin Angizi}, \bibinfo{person}{Zhezhi
  He}, \bibinfo{person}{Adnan~Siraj Rakin}, {and} \bibinfo{person}{Deliang
  Fan}.} \bibinfo{year}{2018}\natexlab{}.
\newblock \showarticletitle{CMP-PIM: An Energy-Efficient Comparator-Based
  Processing-in-Memory Neural Network Accelerator}. In
  \bibinfo{booktitle}{\emph{Proceedings of the 55th Annual Design Automation
  Conference}} (San Francisco, California) \emph{(\bibinfo{series}{DAC '18})}.
  \bibinfo{publisher}{Association for Computing Machinery},
  \bibinfo{address}{New York, NY, USA}, Article \bibinfo{articleno}{105},
  \bibinfo{numpages}{6}~pages.
\newblock
\showISBNx{9781450357005}
\urldef\tempurl%
\url{https://doi.org/10.1145/3195970.3196009}
\showDOI{\tempurl}


\bibitem[Banbury et~al\mbox{.}(2021)]%
        {mlperftiny}
\bibfield{author}{\bibinfo{person}{Colby Banbury},
  \bibinfo{person}{Vijay~Janapa Reddi}, \bibinfo{person}{Peter Torelli},
  \bibinfo{person}{Jeremy Holleman}, \bibinfo{person}{Nat Jeffries},
  \bibinfo{person}{Csaba Kiraly}, \bibinfo{person}{Pietro Montino},
  \bibinfo{person}{David Kanter}, \bibinfo{person}{Sebastian Ahmed},
  \bibinfo{person}{Danilo Pau}, {et~al\mbox{.}}}
  \bibinfo{year}{2021}\natexlab{}.
\newblock \showarticletitle{Mlperf tiny benchmark}.
\newblock \bibinfo{journal}{\emph{arXiv preprint arXiv:2106.07597}}
  (\bibinfo{year}{2021}).
\newblock


\bibitem[Chen et~al\mbox{.}(2016)]%
        {Chen2016Gradient}
\bibfield{author}{\bibinfo{person}{Tianqi Chen}, \bibinfo{person}{Bing Xu},
  \bibinfo{person}{Chiyuan Zhang}, {and} \bibinfo{person}{Carlos Guestrin}.}
  \bibinfo{year}{2016}\natexlab{}.
\newblock \showarticletitle{Training Deep Nets with Sublinear Memory Cost}.
\newblock \bibinfo{journal}{\emph{CoRR}}  \bibinfo{volume}{abs/1604.06174}
  (\bibinfo{year}{2016}).
\newblock
\showeprint[arXiv]{1604.06174}
\urldef\tempurl%
\url{http://arxiv.org/abs/1604.06174}
\showURL{%
\tempurl}


\bibitem[Chetlur et~al\mbox{.}(2014)]%
        {nvidia2014cudnn}
\bibfield{author}{\bibinfo{person}{Sharan Chetlur}, \bibinfo{person}{Cliff
  Woolley}, \bibinfo{person}{Philippe Vandermersch}, \bibinfo{person}{Jonathan
  Cohen}, \bibinfo{person}{John Tran}, \bibinfo{person}{Bryan Catanzaro}, {and}
  \bibinfo{person}{Evan Shelhamer}.} \bibinfo{year}{2014}\natexlab{}.
\newblock \showarticletitle{cuDNN: Efficient Primitives for Deep Learning}.
\newblock \bibinfo{journal}{\emph{CoRR}}  \bibinfo{volume}{abs/1410.0759}
  (\bibinfo{year}{2014}).
\newblock
\showeprint[arXiv]{1410.0759}
\urldef\tempurl%
\url{http://arxiv.org/abs/1410.0759}
\showURL{%
\tempurl}


\bibitem[Gong et~al\mbox{.}(2022)]%
        {Jing2022Approx}
\bibfield{author}{\bibinfo{person}{Jing Gong}, \bibinfo{person}{Hassaan
  Saadat}, \bibinfo{person}{Hasindu Gamaarachchi}, \bibinfo{person}{Haris
  Javaid}, \bibinfo{person}{Xiaobo~Sharon Hu}, {and} \bibinfo{person}{Sri
  Parameswaran}.} \bibinfo{year}{2022}\natexlab{}.
\newblock \bibinfo{title}{ApproxTrain: Fast Simulation of Approximate
  Multipliers for DNN Training and Inference}.
\newblock
\newblock
\urldef\tempurl%
\url{https://doi.org/10.48550/ARXIV.2209.04161}
\showDOI{\tempurl}


\bibitem[He et~al\mbox{.}(2016)]%
        {resnet}
\bibfield{author}{\bibinfo{person}{Kaiming He}, \bibinfo{person}{Xiangyu
  Zhang}, \bibinfo{person}{Shaoqing Ren}, {and} \bibinfo{person}{Jian Sun}.}
  \bibinfo{year}{2016}\natexlab{}.
\newblock \showarticletitle{Deep residual learning for image recognition}. In
  \bibinfo{booktitle}{\emph{Proceedings of the IEEE conference on computer
  vision and pattern recognition}}. \bibinfo{pages}{770--778}.
\newblock


\bibitem[Huang et~al\mbox{.}(2021)]%
        {cimquantization1}
\bibfield{author}{\bibinfo{person}{Sitao Huang}, \bibinfo{person}{Aayush
  Ankit}, \bibinfo{person}{Plinio Silveira}, \bibinfo{person}{Rodrigo Antunes},
  \bibinfo{person}{Sai~Rahul Chalamalasetti}, \bibinfo{person}{Izzat El~Hajj},
  \bibinfo{person}{Dong~Eun Kim}, \bibinfo{person}{Glaucimar Aguiar},
  \bibinfo{person}{Pedro Bruel}, \bibinfo{person}{Sergey Serebryakov},
  \bibinfo{person}{Cong Xu}, \bibinfo{person}{Can Li}, \bibinfo{person}{Paolo
  Faraboschi}, \bibinfo{person}{John~Paul Strachan}, \bibinfo{person}{Deming
  Chen}, \bibinfo{person}{Kaushik Roy}, \bibinfo{person}{Wen-mei Hwu}, {and}
  \bibinfo{person}{Dejan Milojicic}.} \bibinfo{year}{2021}\natexlab{}.
\newblock \showarticletitle{Mixed Precision Quantization for ReRAM-based DNN
  Inference Accelerators}. In \bibinfo{booktitle}{\emph{2021 26th Asia and
  South Pacific Design Automation Conference (ASP-DAC)}}.
  \bibinfo{pages}{372--377}.
\newblock


\bibitem[Kirtas et~al\mbox{.}(2022)]%
        {photonicquantization1}
\bibfield{author}{\bibinfo{person}{M. Kirtas}, \bibinfo{person}{A. Oikonomou},
  \bibinfo{person}{N. Passalis}, \bibinfo{person}{G. Mourgias-Alexandris},
  \bibinfo{person}{M. Moralis-Pegios}, \bibinfo{person}{N. Pleros}, {and}
  \bibinfo{person}{A. Tefas}.} \bibinfo{year}{2022}\natexlab{}.
\newblock \showarticletitle{Quantization-aware training for low precision
  photonic neural networks}.
\newblock \bibinfo{journal}{\emph{Neural Networks}}  \bibinfo{volume}{155}
  (\bibinfo{year}{2022}), \bibinfo{pages}{561--573}.
\newblock
\showISSN{0893-6080}
\urldef\tempurl%
\url{https://doi.org/10.1016/j.neunet.2022.09.015}
\showDOI{\tempurl}


\bibitem[Kulkarni et~al\mbox{.}(2011)]%
        {Kulkarni2011Approx}
\bibfield{author}{\bibinfo{person}{Parag Kulkarni}, \bibinfo{person}{Puneet
  Gupta}, {and} \bibinfo{person}{Milos Ercegovac}.}
  \bibinfo{year}{2011}\natexlab{}.
\newblock \showarticletitle{Trading Accuracy for Power with an Underdesigned
  Multiplier Architecture}. In \bibinfo{booktitle}{\emph{2011 24th Internatioal
  Conference on VLSI Design}}. \bibinfo{pages}{346--351}.
\newblock
\urldef\tempurl%
\url{https://doi.org/10.1109/VLSID.2011.51}
\showDOI{\tempurl}


\bibitem[Lai et~al\mbox{.}(2018)]%
        {Lai2018MlCmsis}
\bibfield{author}{\bibinfo{person}{Liangzhen Lai}, \bibinfo{person}{Naveen
  Suda}, {and} \bibinfo{person}{Vikas Chandra}.}
  \bibinfo{year}{2018}\natexlab{}.
\newblock \showarticletitle{{CMSIS}-{NN}: {Efficient} {Neural} {Network}
  {Kernels} for {Arm} {Cortex}-{M} {CPUs}}.
\newblock  (\bibinfo{year}{2018}), \bibinfo{pages}{1--10}.
\newblock
\urldef\tempurl%
\url{http://arxiv.org/abs/1801.06601}
\showURL{%
\tempurl}
\newblock
\shownote{arXiv: 1801.06601}.


\bibitem[Lee et~al\mbox{.}(2021)]%
        {drampim}
\bibfield{author}{\bibinfo{person}{Sukhan Lee}, \bibinfo{person}{Shin-haeng
  Kang}, \bibinfo{person}{Jaehoon Lee}, \bibinfo{person}{Hyeonsu Kim},
  \bibinfo{person}{Eojin Lee}, \bibinfo{person}{Seungwoo Seo},
  \bibinfo{person}{Hosang Yoon}, \bibinfo{person}{Seungwon Lee},
  \bibinfo{person}{Kyounghwan Lim}, \bibinfo{person}{Hyunsung Shin},
  {et~al\mbox{.}}} \bibinfo{year}{2021}\natexlab{}.
\newblock \showarticletitle{Hardware architecture and software stack for PIM
  based on commercial DRAM technology: Industrial product}. In
  \bibinfo{booktitle}{\emph{2021 ACM/IEEE 48th Annual International Symposium
  on Computer Architecture (ISCA)}}. IEEE, \bibinfo{pages}{43--56}.
\newblock


\bibitem[Li et~al\mbox{.}(2022)]%
        {li2022photofourier}
\bibfield{author}{\bibinfo{person}{Shurui Li}, \bibinfo{person}{Hangbo Yang},
  \bibinfo{person}{Chee~Wei Wong}, \bibinfo{person}{Volker~J Sorger}, {and}
  \bibinfo{person}{Puneet Gupta}.} \bibinfo{year}{2022}\natexlab{}.
\newblock \showarticletitle{PhotoFourier: A Photonic Joint Transform
  Correlator-Based Neural Network Accelerator}.
\newblock \bibinfo{journal}{\emph{arXiv preprint arXiv:2211.05276}}
  (\bibinfo{year}{2022}).
\newblock


\bibitem[Li et~al\mbox{.}(2021)]%
        {Li2021Geo}
\bibfield{author}{\bibinfo{person}{Tianmu Li}, \bibinfo{person}{Wojciech
  Romaszkan}, \bibinfo{person}{Sudhakar Pamarti}, {and} \bibinfo{person}{Puneet
  Gupta}.} \bibinfo{year}{2021}\natexlab{}.
\newblock \showarticletitle{{GEO} : {Generation} and {Execution} {Optimized}
  {Stochastic} {Computing} {Accelerator} for {Neural} {Networks}}. In
  \bibinfo{booktitle}{\emph{2021 {Design}, {Automation} \& {Test} in {Europe}
  {Conference} \& {Exhibition} ({DATE})}}. \bibinfo{pages}{1--6}.
\newblock


\bibitem[Mrazek et~al\mbox{.}(2017)]%
        {Mrazek2017Evo}
\bibfield{author}{\bibinfo{person}{Vojtech Mrazek}, \bibinfo{person}{Radek
  Hrbacek}, \bibinfo{person}{Zdenek Vasicek}, {and} \bibinfo{person}{Lukas
  Sekanina}.} \bibinfo{year}{2017}\natexlab{}.
\newblock \showarticletitle{EvoApprox8b: Library of Approximate Adders and
  Multipliers for Circuit Design and Benchmarking of Approximation Methods}. In
  \bibinfo{booktitle}{\emph{Design, Automation \& Test in Europe Conference \&
  Exhibition (DATE), 2017}}. \bibinfo{pages}{258--261}.
\newblock
\urldef\tempurl%
\url{https://doi.org/10.23919/DATE.2017.7926993}
\showDOI{\tempurl}


\bibitem[Oikonomou et~al\mbox{.}(2022)]%
        {photonicquantization2}
\bibfield{author}{\bibinfo{person}{A. Oikonomou}, \bibinfo{person}{M. Kirtas},
  \bibinfo{person}{N. Passalis}, \bibinfo{person}{G. Mourgias-Alexandris},
  \bibinfo{person}{M. Moralis-Pegios}, \bibinfo{person}{N. Pleros}, {and}
  \bibinfo{person}{A. Tefas}.} \bibinfo{year}{2022}\natexlab{}.
\newblock \showarticletitle{A Robust, Quantization-Aware Training Method
  for Photonic Neural Networks}. In \bibinfo{booktitle}{\emph{Engineering
  Applications of Neural Networks}}, \bibfield{editor}{\bibinfo{person}{Lazaros
  Iliadis}, \bibinfo{person}{Chrisina Jayne}, \bibinfo{person}{Anastasios
  Tefas}, {and} \bibinfo{person}{Elias Pimenidis}} (Eds.).
  \bibinfo{publisher}{Springer International Publishing},
  \bibinfo{address}{Cham}, \bibinfo{pages}{427--438}.
\newblock


\bibitem[Rastegari et~al\mbox{.}(2016)]%
        {Rastegari2016XNOR}
\bibfield{author}{\bibinfo{person}{Mohammad Rastegari},
  \bibinfo{person}{Vicente Ordonez}, \bibinfo{person}{Joseph Redmon}, {and}
  \bibinfo{person}{Ali Farhadi}.} \bibinfo{year}{2016}\natexlab{}.
\newblock \showarticletitle{XNOR-Net: ImageNet Classification Using Binary
  Convolutional Neural Networks}.
\newblock \bibinfo{journal}{\emph{CoRR}}  \bibinfo{volume}{abs/1603.05279}
  (\bibinfo{year}{2016}).
\newblock
\showeprint[arXiv]{1603.05279}
\urldef\tempurl%
\url{http://arxiv.org/abs/1603.05279}
\showURL{%
\tempurl}


\bibitem[Romaszkan et~al\mbox{.}(2020)]%
        {Romaszkan2020Acoustic}
\bibfield{author}{\bibinfo{person}{Wojciech Romaszkan}, \bibinfo{person}{Tianmu
  Li}, \bibinfo{person}{Tristan Melton}, \bibinfo{person}{Sudhakar Pamarti},
  {and} \bibinfo{person}{Puneet Gupta}.} \bibinfo{year}{2020}\natexlab{}.
\newblock \showarticletitle{ACOUSTIC: Accelerating Convolutional Neural
  Networks through Or-Unipolar Skipped Stochastic Computing}. In
  \bibinfo{booktitle}{\emph{2020 Design, Automation \& Test in Europe
  Conference \& Exhibition (DATE)}}. \bibinfo{pages}{768--773}.
\newblock
\urldef\tempurl%
\url{https://doi.org/10.23919/DATE48585.2020.9116289}
\showDOI{\tempurl}


\bibitem[Shiflett et~al\mbox{.}(2021)]%
        {shiflett2021albireo}
\bibfield{author}{\bibinfo{person}{Kyle Shiflett}, \bibinfo{person}{Avinash
  Karanth}, \bibinfo{person}{Razvan Bunescu}, {and} \bibinfo{person}{Ahmed
  Louri}.} \bibinfo{year}{2021}\natexlab{}.
\newblock \showarticletitle{Albireo: Energy-efficient acceleration of
  convolutional neural networks via silicon photonics}. In
  \bibinfo{booktitle}{\emph{2021 ACM/IEEE 48th Annual International Symposium
  on Computer Architecture (ISCA)}}. IEEE, \bibinfo{pages}{860--873}.
\newblock


\bibitem[Sim and Lee(2017)]%
        {Sim2017BISC}
\bibfield{author}{\bibinfo{person}{Hyeonuk Sim} {and} \bibinfo{person}{Jongeun
  Lee}.} \bibinfo{year}{2017}\natexlab{}.
\newblock \showarticletitle{A new stochastic computing multiplier with
  application to deep convolutional neural networks}. In
  \bibinfo{booktitle}{\emph{2017 54th ACM/EDAC/IEEE Design Automation
  Conference (DAC)}}. \bibinfo{pages}{1--6}.
\newblock
\urldef\tempurl%
\url{https://doi.org/10.1145/3061639.3062290}
\showDOI{\tempurl}


\bibitem[Venkataramani et~al\mbox{.}(2014)]%
        {Venkataramani2014Axnn}
\bibfield{author}{\bibinfo{person}{Swagath Venkataramani},
  \bibinfo{person}{Ashish Ranjan}, \bibinfo{person}{Kaushik Roy}, {and}
  \bibinfo{person}{Anand Raghunathan}.} \bibinfo{year}{2014}\natexlab{}.
\newblock \showarticletitle{AxNN: Energy-efficient neuromorphic systems using
  approximate computing}. In \bibinfo{booktitle}{\emph{2014 IEEE/ACM
  International Symposium on Low Power Electronics and Design (ISLPED)}}.
  \bibinfo{pages}{27--32}.
\newblock
\urldef\tempurl%
\url{https://doi.org/10.1145/2627369.2627613}
\showDOI{\tempurl}


\bibitem[Zhang et~al\mbox{.}(2015)]%
        {Zhang2015Approx}
\bibfield{author}{\bibinfo{person}{Qian Zhang}, \bibinfo{person}{Ting Wang},
  \bibinfo{person}{Ye Tian}, \bibinfo{person}{Feng Yuan}, {and}
  \bibinfo{person}{Qiang Xu}.} \bibinfo{year}{2015}\natexlab{}.
\newblock \showarticletitle{ApproxANN: An approximate computing framework for
  artificial neural network}. In \bibinfo{booktitle}{\emph{2015 Design,
  Automation \& Test in Europe Conference \& Exhibition (DATE)}}.
  \bibinfo{pages}{701--706}.
\newblock


\bibitem[Zhang et~al\mbox{.}(2019)]%
        {cimquantization2}
\bibfield{author}{\bibinfo{person}{Wenqiang Zhang}, \bibinfo{person}{Xiaochen
  Peng}, \bibinfo{person}{Huaqiang Wu}, \bibinfo{person}{Bin Gao},
  \bibinfo{person}{Hu He}, \bibinfo{person}{Youhui Zhang},
  \bibinfo{person}{Shimeng Yu}, {and} \bibinfo{person}{He Qian}.}
  \bibinfo{year}{2019}\natexlab{}.
\newblock \showarticletitle{Design Guidelines of RRAM based
  Neural-Processing-Unit: A Joint Device-Circuit-Algorithm Analysis}. In
  \bibinfo{booktitle}{\emph{2019 56th ACM/IEEE Design Automation Conference
  (DAC)}}. \bibinfo{pages}{1--6}.
\newblock


\end{thebibliography}

\end{document}